\documentclass[10pt,twocolumn,letterpaper]{article}

\usepackage{cvpr}
\usepackage{times}
\usepackage{epsfig}
\usepackage{graphicx}
\usepackage{amsmath}
\usepackage{amssymb}
\usepackage{xcolor}
\usepackage{tikz}

\definecolor{yifei}{RGB}{0, 0, 0}
\definecolor{liu}{RGB}{0, 0, 0}
\definecolor{zirui}{RGB}{0, 0, 0}

\cvprfinalcopy
\def\cvprPaperID{****}

\begin{document}

\title{\huge{In-Context Learning Distillation for Efficient Few-Shot Fine-Tuning}}

\author{\textbf{Yifei DUAN,~ Liu LI,~ Zirui ZHAI,~ Jinxia YAO
}\\
\normalsize{Georgia Institute of Technology}
\\
{\tt\small {yduan92,lli665,zzhai38,jyao320}@gatech.edu}
}

\maketitle

\begin{abstract}
\textcolor{yifei}{
We applied few-shot in-context learning on the OPT-1.3B model for the natural language inference task and employed knowledge distillation to internalize the context information, reducing model parameter from 1.3B to 125M and achieving a size reduction from 2.5GB to 0.25GB. Compared to using in-context learning alone on similarly sized models, this context distillation approach achieved a nearly 50\% improvement in out-of-domain accuracy, demonstrating superior knowledge transfer capabilities over prompt-based methods. Furthermore, this approach reduced memory consumption by up to 60\% while delivering a 20\% improvement in out-of-domain accuracy compared to conventional pattern-based fine-tuning.
}
\end{abstract}

\section{Introduction/Background/Motivation}

Large Language Models (LLMs) have revolutionized natural language processing, demonstrating remarkable capabilities in various tasks. However, their deployment faces significant challenges, particularly with regard to memory requirements during training and inference. In-context learning (ICL)\cite{mosbach2023few}, while effective, requires substantial computational resources due to the need to maintain long context windows. This limitation becomes particularly severe when dealing with tasks that require extensive context processing, such as long-form dialogue or document analysis. The core challenge lies in the trade-off between model performance and resource efficiency. Current approaches require extensive computational resources (full fine-tuning)\cite{lv2023full} or significant memory overhead (ICL). This creates a barrier to training, inference and deployment in resource-constrained environments and limits the practical applicability of LLMs in real-world scenarios.
  
\textcolor{liu}{
Few-shot learning is used for task adaptation by enabling the model to generalize to specific tasks with minimal labeled data. This approach leverages a small number of training examples, which makes it particularly valuable for scenarios where obtaining large-scale annotated datasets is impractical.
Conventional solutions to few-shot learning generally fall into two categories: weights-updating fine-tuning and prompt-based context learning.}
\textcolor{liu}{Each approach has significant limitations, particularly when scaling to larger models or deploying in resource-constrained environments. Fine-tuning requires updating some or all model parameters, leading to high computational costs and potential catastrophic forgetting.}
\textcolor{liu}{In-Context Learning demands large memory allocations for context windows, limiting scalability.}


\textcolor{liu}{In this paper, we implemented Context Distillation (CD)\cite{snell2022learning}, a novel approach that combines the benefits of fine-tuning and in-context learning while minimizing their respective drawbacks. Our key contributions include:} \textcolor{liu}{(1) A novel context distillation methodology that efficiently captures prompt information in model parameters.}
\textcolor{liu}{(2) Integration with parameter-efficient fine-tuning techniques (BitFit\cite{zaken2021bitfit} and LoRA\cite{hu2021lora}).}
\textcolor{liu}{(3) Comprehensive experiment designs to study the impact of critical hyper-parameters and model sizes horizontally and vertically.} 
This approach provides a resource-efficient solution that balances performance with hardware constraints, enabling the deployment of advanced LLM capabilities, particularly for on-device applications.

\textcolor{liu}{The Multi-Genre Natural Language Inference (MNLI) corpus \cite{williams2017broad} is used for all training and evaluation in this paper. It contains sentence pairs labeled as entailment, contradiction, or neutral, designed to train and assess models on natural language inference tasks. The dataset includes separate training, development, and testing sets, divided into matched (in-domain) and mismatched (cross-domain) subsets to evaluate generalization. Each data point consists of a "premise" sentence, a "hypothesis" sentence, and a label indicating their relationship.}

\section{Approach}
\subsection{implementation}

\textcolor{yifei}{
The hardware setup for this experiment includes NVIDIA T4, A100, and RTX 3090 GPUs. The T4 GPU, with at least 16 GB of RAM, is sufficient for performing inference on models up to 1.3 billion parameters in half precision, supporting up to 12 support examples per evaluation and a training size of 20 in this study. 
The implementation is developed using PyTorch and Python 3.9, primarily on the RTX 3090 GPU. Most evaluations and analyses are conducted on cloud-based A100 GPUs to ensure consistency in training time and memory usage, eliminating hardware variability as a factor in the results.
}

\textcolor{yifei}{
The MNLI dataset is first refined into a binary classification task by removing instances labeled as neutral. For efficient processing, the dataset is reduced to a pool of 3,000 examples, from which support cases are randomly selected while ensuring the query case is excluded from the support set. 
Experiments show that varying the pool size to 1,000 or 10,000 examples has negligible impact on the results, demonstrating the robustness of the approach to pool size variations.
This is expected as the few-shot training data is limited to 20 examples, regardless of the size of the pool they are drawn from. 
The methods ICL, PBFT, and CD require prompts structured with three components: prefix, support, and query. The support component includes 1 to 12 examples. A sample prompt structure is provided in Table \ref{tab:prompt_prediction}.
}

\textcolor{yifei}{
Few-shot fine-tuning leverages contextual embeddings during the model training process, employing the AdamW optimizer with a learning rate of 1e-5 and a 40-epoch schedule. Evaluations are performed using in-domain and out-of-domain accuracy, along with peak memory usage. GPU memory utilization is tracked with utility tools such as `torch\_peak\_allocated\_memory', and computational efficiency is measured using `timer' over 100 cases per evaluation. While `timer' may not fully account for asynchronous GPU operations, its error margin of less than 1 second is deemed acceptable for this study. 
Results are benchmarked against the pre-trained model without fine-tuning, providing insights into the effectiveness and efficiency of context distillation in enhancing model performance.
}

\textcolor{yifei}{
To adapt the large language model (LLM) for binary classification, a verbalizer was utilized to map the model’s output logits to interpretable labels, specifically using "yes" for entailment and "no" for contradiction. This simplified label space enhanced decision-making while leveraging the pre-trained model’s ability to assign meaningful probabilities to natural language tokens. During evaluation, the verbalizer mapping was consistently applied across both in-domain (MNLI-matched) and out-of-domain (MNLI-mismatched) validation splits, enabling a comprehensive assessment of model performance. 
\begin{table}[h!]
\centering
\small
\setlength{\tabcolsep}{5pt}
\renewcommand{\arraystretch}{1.2}
\begin{tabular}{|p{2cm}|p{5cm}|}
\hline
\textbf{Label} & \textbf{Content} \\ \hline
 Prefix & Determine if the premise entails the hypothesis. Answer with yes or no.  \\ \hline
Support & Premise: It was a steep learning curve for me, she said. \\ 
               & Hypothesis: She faced no difficulty with the task. \\ 
               & Answer: No \\ \hline
Query     & Premise: I'll listen and agree with what I think sounds right. \\ 
               & Hypothesis: I won't even bother listening. \\ 
               & Answer: \\ \hline
Prediction & Yes (based on the logit outputs of verbalizers) \\ \hline
\end{tabular}
\caption{Sample prompt with one support  and one query  for pattern-based fine-tuning and context distillation. The table reproduces the structure of the prompt used in the experiment.}
\label{tab:prompt_prediction}
\end{table}
}

\textcolor{black}{
To ensure efficient and consistent fine-tuning, we independently implemented two parameter-efficient techniques: LoRA and BitFit. LoRA used a low-rank matrix rank (\texttt{r}) of 8, a scaling factor (\texttt{lora\_alpha}) of 32, and targeted the \texttt{q\_proj} and \texttt{v\_proj} layers with a dropout rate (\texttt{lora\_dropout}) of 0.05. This setup was optimized for sequence classification tasks (\texttt{task\_type="SEQ\_CLS"}) and allowed approximately 23\% of the model’s parameters to remain trainable.
BitFit, on the other hand, fine-tuned only the bias parameters, freezing all other weights. 
In the investigated cases, less than 0.1\% of the parameters are trainable for BitFit.
}

\subsection{Few-shot fine-tuning overfitting}
\textcolor{yifei}{
Few-shot fine-tuning is particularly prone to overfitting, as observed in our experiments, where increasing the number of epochs leads to a decline in out-of-domain accuracy. To counter this, hyperparameters such as learning rate and batch size were carefully optimized to ensure controlled parameter updates, reducing the model's susceptibility to noise in the limited data. Learning rates were tested systematically within the range of 1e-4 to 1e-7, while the number of epochs was varied between 2 and 100. The optimal configuration of these hyperparameters was selected based on maximizing out-of-domain accuracy, striking a balance between effective learning and prevention of overfitting. Moreover, parameter-efficient fine-tuning (PEFT) techniques, such as LoRA and BitFit, were employed to leverage pre-trained knowledge by freezing or significantly reducing the number of trainable parameters. These techniques demonstrated improved accuracy compared to traditional fine-tuning approaches without PEFT methods. By preserving general representations learned from larger datasets while allowing task-specific adaptation, this approach effectively mitigates overfitting. 
}

\subsection{Uncertainties in accuracy evaluation}

\textcolor{yifei}{
To ensure that the measured accuracy reliably reflects the model’s performance, evaluations were performed using inferences ranging from 100 to 1,000. For each inference count, five runs were conducted with different random seeds, and the final accuracy results were averaged over these runs. The results reveal that the number of inferences has a minimal impact on accuracy, with deviations of approximately $\pm 15\%$. For efficient evaluation purposes, accuracy is determined from 100 inferences, recognizing that each inference may include multiple support cases, particularly for methods such as in-context learning. Additionally, since the accuracy results remain consistent across runs with different random seeds, a fixed random seed is utilized in subsequent analyses to ensure reproducibility and consistency across runs and platforms. This setup effectively balances computational efficiency with robust and reliable evaluation, ensuring fair comparisons of model performance.
}

\subsection{Loss imbalance in context distillation}
\textcolor{yifei}{
For training processes that involve combining multiple loss functions, it is essential to ensure that the losses are balanced to prevent one from dominating the total loss. In our approach, the loss is primarily based on the KL-divergence between the student and teacher logits. To test this, we employed a dynamic balance between the KL-divergence loss ($L_{KL}$) and the cross-entropy loss ($L_{ce}$) between the student’s predictions and the true labels. This is represented by the equation:
\begin{equation}
L=\alpha L_{KL}+ (1-\alpha)L_{ce}
\end{equation}
where $\alpha$ is a balancing factor. When $\alpha = 0$, the context distillation reduces to vanilla fine-tuning, focusing solely on the cross-entropy loss. Conversely, when $\alpha = 1$, the approach becomes equivalent to knowledge distillation, relying entirely on the KL-divergence loss. By varying $\alpha$, we aim to find an optimal balance that maximizes the performance of the student model, effectively leveraging both the teacher’s knowledge and the true labels to achieve robust learning. Our experiments reveal that out-of-domain accuracy initially increases with $\alpha$ but subsequently decreases beyond a certain point. Based on these observations, we determined that $\alpha=0.5$ provides the best trade-off, yielding optimal performance by harmonizing the contributions of the teacher’s predictions and the true labels.
}

\subsection{Selection of Verbalizers}

\textcolor{yifei}{
The selection of verbalizers plays a critical role in determining the performance of a model, particularly in binary classification tasks. To create a concise verbalizer pool, we designed prompts that explicitly instructed the model to respond with “Yes” or “No,” as demonstrated in Table \ref{tab:prompt_prediction}. This strategy ensured that the model’s top two token predictions were “Yes” and “No,” which collectively accounted for over 40\% of the total probability. During experimentation, we noticed that verbalizers starting with lowercase, such as “yes,” could occasionally be tokenized differently. To address this, we combined “Yes” and “yes” as valid verbalizers, although results indicated that including “yes” did not alter performance. This observation underscores the robustness of this verbalizer choice, providing reliable outcomes while minimizing ambiguity in binary classification tasks.
}

\subsection{LLM with half precision}
\textcolor{yifei}{
During our experiment with pattern-based fine tuning, we observed that fine-tuning required a significant amount of memory. Specifically, context distillation proved infeasible on a 24 GB GPU when using a teacher model of 1.3 billion parameters and a student model of 125 million parameters. To address this, we imported the model in FP16 precision, which substantially reduced the memory footprint and enabled us to deploy the 1.3B teacher model. However, we encountered a critical issue: FP16 precision resulted in loss values being returned as ‘NaN,’ likely due to its insufficient dynamic range for our setup. To resolve this, we switched to BF16 precision, which maintains the same dynamic range as FP32 but sacrifices some accuracy. This adjustment proved effective, as BF16 accommodated our memory constraints while avoiding the loss stability issues encountered with FP16. This approach allowed for successful execution of context distillation while significantly reducing memory usage.
}

\section{Experiments and Results}

\textcolor{zirui}{
\subsection{Experiments setup}
We conducted a comprehensive set of controlled experiments to systematically evaluate and compare multiple approaches, including the baseline method (pre-trained model without task adaption), In-Context Learning (ICL), Parameter-Based Fine-Tuning (PBFT), and Context Distillation (CD) techniques, along with their parameter-efficient variants (e.g., PBFT+LoRA, PBFT+BitFit, CD+LoRA, CD+BitFit). In these experiments, we varied the number of support examples (e.g., 1, 4, 8, 12) and recorded in-domain and out-of-domain accuracy, peak memory usage, and total training and evaluation times. By applying this consistent setup across all methods, we were able to directly assess their performance stability, resource consumption, and scalability, thereby identifying key trade-offs and guiding principles for selecting the most suitable method under different practical constraints. For further details on the parameter settings and experimental configurations, please refer to Section 2.1.
}

\begin{figure}[h]
\begin{center}
   \includegraphics[width=\linewidth]{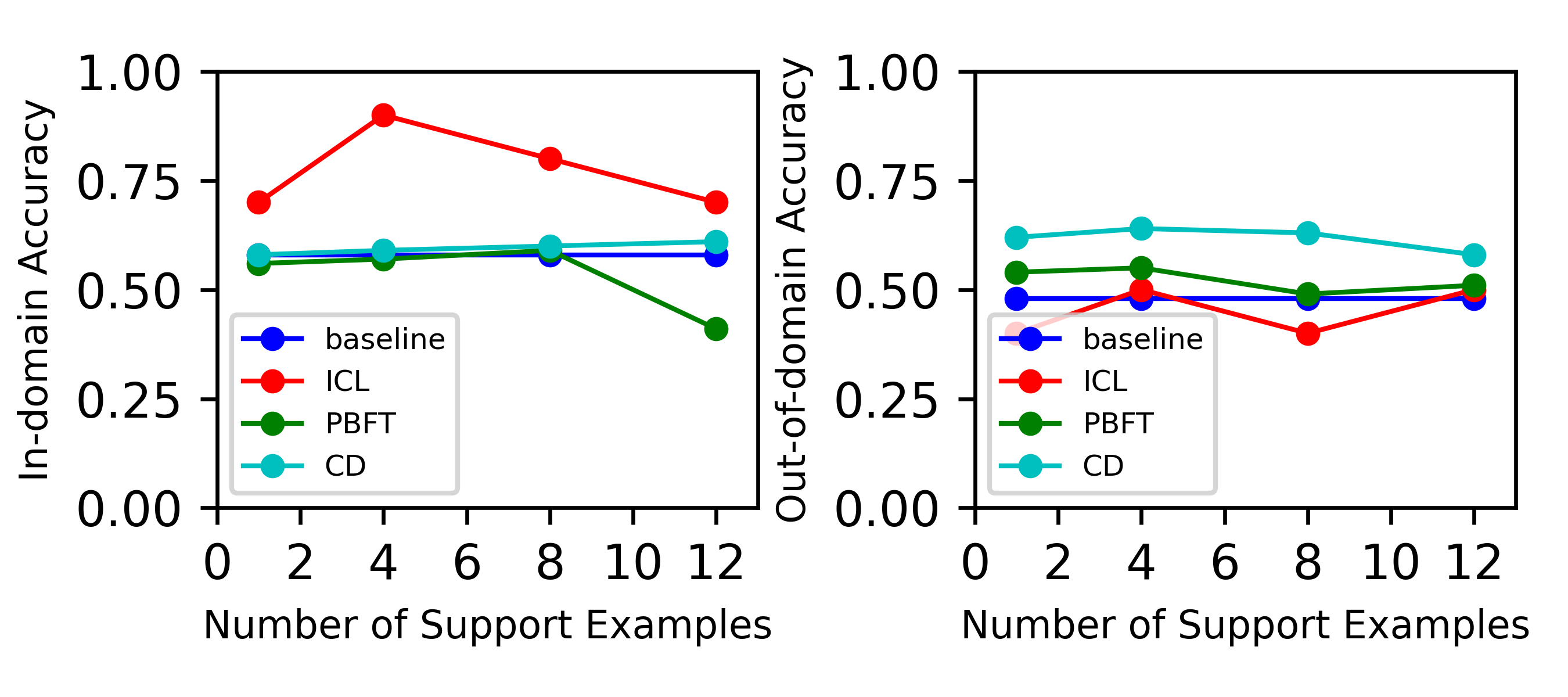}
\end{center}
   \caption{\textcolor{yifei}{In-domain and out-of-domain accuracy for different task adaptation techniques applied to OPT-125M model vs.  numbers of support examples.}}
   \label{fig:125m-baseline-acc}
\end{figure}

\textcolor{zirui}{
\subsection{Baseline, ICL, PBFT, Context Distillation}
Our experimental results (Figure \ref{fig:125m-baseline-acc}) demonstrate distinct patterns in both in-domain and out-of-domain accuracy across the four methods evaluated. In-domain performance shows that ICL (In-Context Learning) consistently outperforms other approaches, achieving peak accuracy of approximately 0.9 with 4 support examples. This represents a significant improvement over the baseline, which maintains relatively stable performance around 0.6 across different numbers of support examples. The Context Distillation (CD) method shows comparable performance to the baseline, while PBFT exhibits declining performance with increased support examples, particularly after 8 examples where accuracy drops to approximately 0.4.
For out-of-domain generalization, the performance dynamics shift notably. CD demonstrates the most robust performance, maintaining accuracy around 0.6-0.65 across different support example quantities. Both ICL and PBFT show more volatile performance patterns, with accuracies fluctuating between 0.4-0.5, suggesting less stable generalization capabilities compared to their in-domain performance.
}

\begin{figure}[h]
\begin{center}
   \includegraphics[width=\linewidth]{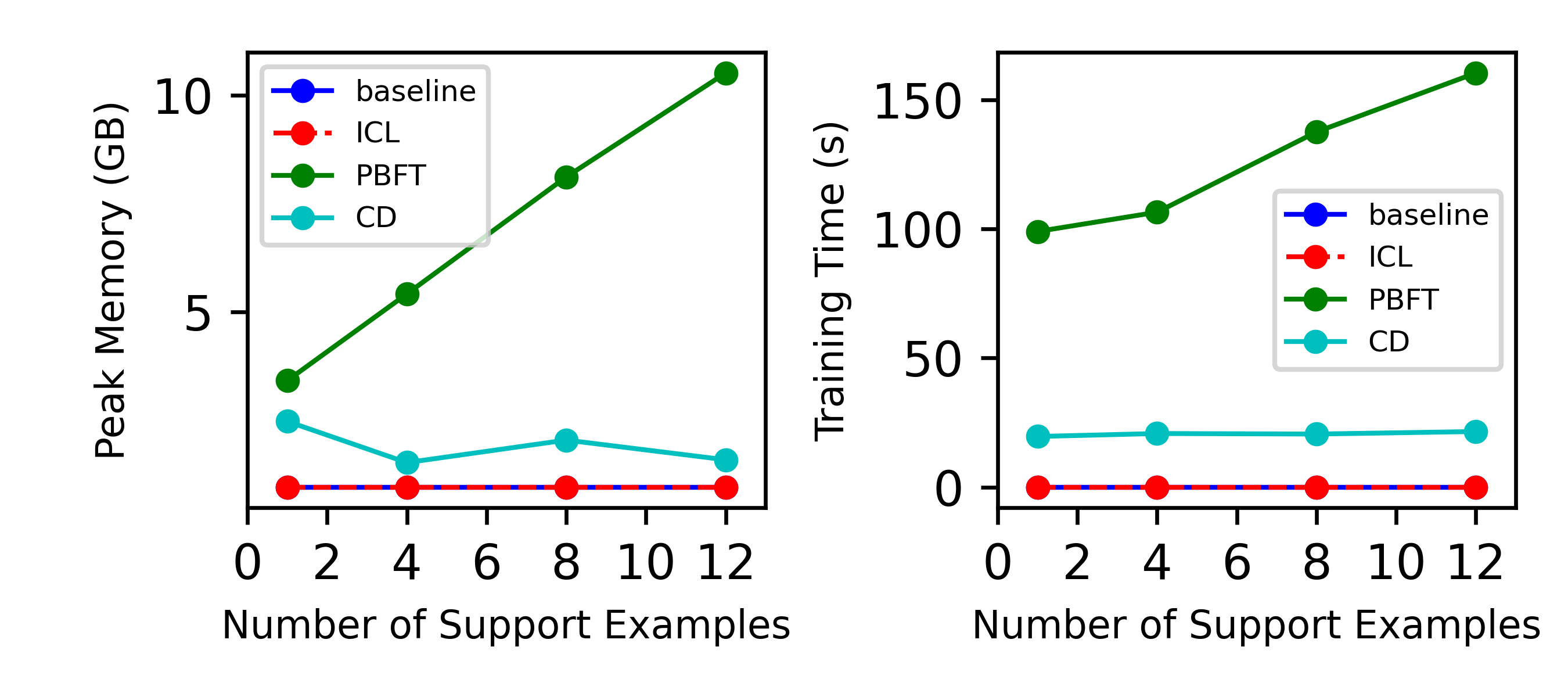}
\end{center}
   \caption{\textcolor{yifei}{Peak GPU memory allocated and training time for different task adaptation techniques applied to OPT-125M model vs. numbers of support examples. Note that the baseline and ICL do not involve fine-tuning; their memory usage reflects only the inference process, and the training time is plotted as 0 for reference}}
   \label{fig:125m-baseline-eff}
\end{figure}

\textcolor{zirui}{
Memory consumption patterns reveal significant differences among the methods (Figure \ref{fig:125m-baseline-eff}, left panel). PBFT shows a linear increase in peak memory usage, scaling from approximately 3GB with 1 support example to over 10GB with 12 support examples. This trend indicates potential scalability concerns for larger datasets.
CD demonstrates moderate memory requirements, stabilizing between 1.5-2GB across different support example quantities, offering a balanced compromise between memory efficiency and performance.
Most importantly, the memory usage for CD is nearly independent of the number of support examples.
Lastly, since ICL and the baseline do not involve fine-tuning, their memory metric is based on the inference process, showing consistent memory usage of approximately 1GB regardless of the support example count. This level of memory usage is also expected for the inference process of PBFT and CD.
}

\textcolor{zirui}{
Figure \ref{fig:125m-baseline-eff} on the right panel shows that training time efficiency varies substantially across methods. PBFT exhibits the highest computational overhead, with training time increasing linearly from approximately 100 seconds to 160 seconds as support examples increase. This scaling behavior suggests potential limitations for large-scale applications. Conversely, CD demonstrates remarkable efficiency, maintaining nearly constant training times regardless of the number of support examples. CD consistently requires around 20 seconds per training session.
Note that the baseline and ICL do not involve fine-tuning; their training time is plotted as 0 for reference.
These results suggest that while ICL achieves superior in-domain performance without requiring fine-tuning, CD offers better generalization capabilities with moderate resource requirements. PBFT, despite its flexibility, faces scalability challenges in both memory usage and training time as the number of support examples increases.
}

\begin{figure}[h]
\begin{center}
   \includegraphics[width=\linewidth]{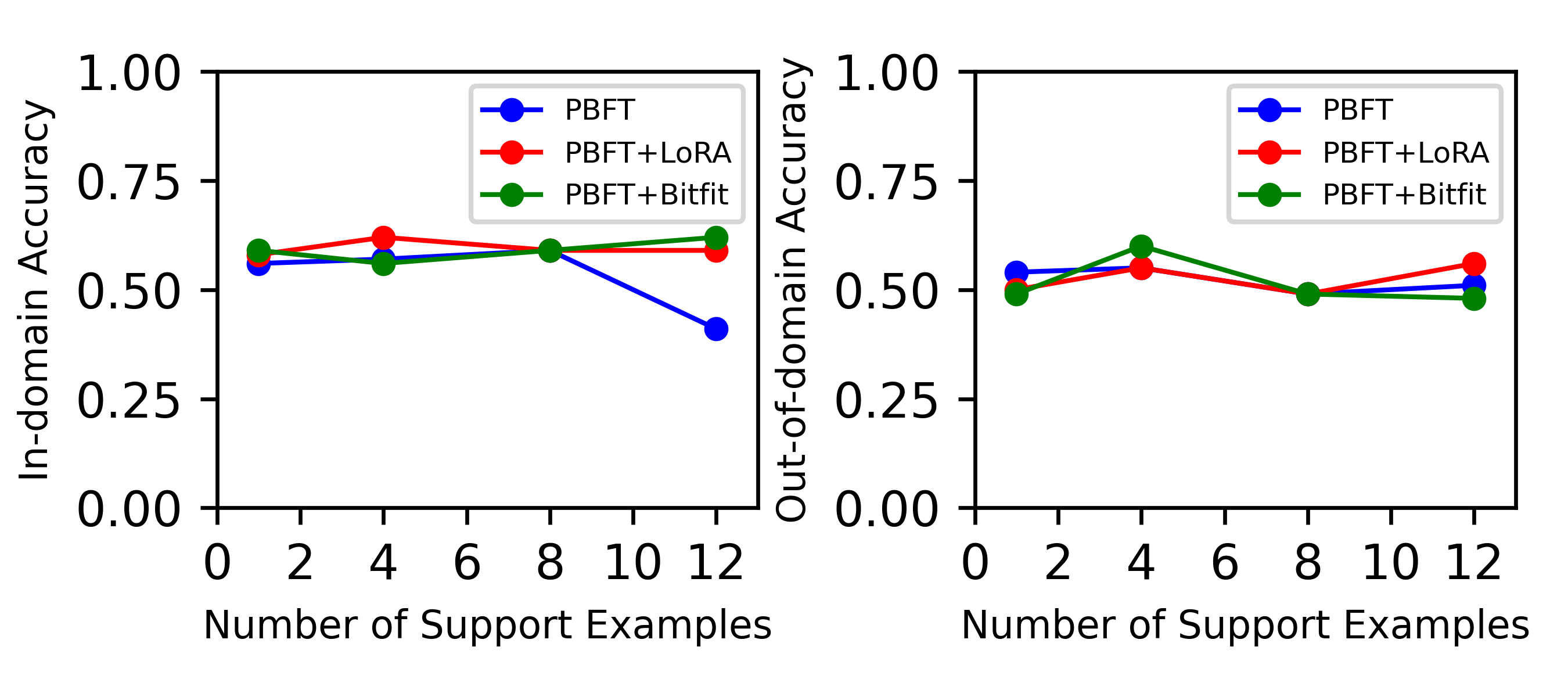}
\end{center}
   \caption{\textcolor{yifei}{In-domain and out-of-domain accuracy for PBFT OPT-125M model using LoRA and BitFit.}}
   \label{fig:125m-pbft-acc}
\end{figure}

\textcolor{zirui}{
\subsection{PBFT+BitFit and PBFT+LoRA Analysis}
Our experimental results demonstrate notable differences between PBFT and its parameter-efficient variants (Figure \ref{fig:125m-pbft-acc}). For in-domain accuracy, both PBFT+BitFit and PBFT+LoRA maintain stronger performance compared to baseline PBFT, particularly with larger numbers of support examples. While standard PBFT’s accuracy decreases notably at 12 support examples (dropping to around 0.4), both variants maintain more stable performance, remaining between approximately 0.55 and 0.62. PBFT+BitFit achieves slightly higher accuracy, peaking at about 0.62 when using 12 support examples, while PBFT+LoRA sustains a consistent range near 0.58–0.62.
In out-of-domain generalization, all three approaches show similar performance patterns, with accuracies ranging between 0.48-0.60. PBFT+BitFit and PBFT+LoRA both demonstrate more stable performance across varying numbers of support examples compared to baseline PBFT, with PBFT+BitFit showing marginally better stability in the 0.5-0.6 range.
}

\begin{figure}[h]
\begin{center}
   \includegraphics[width=\linewidth]{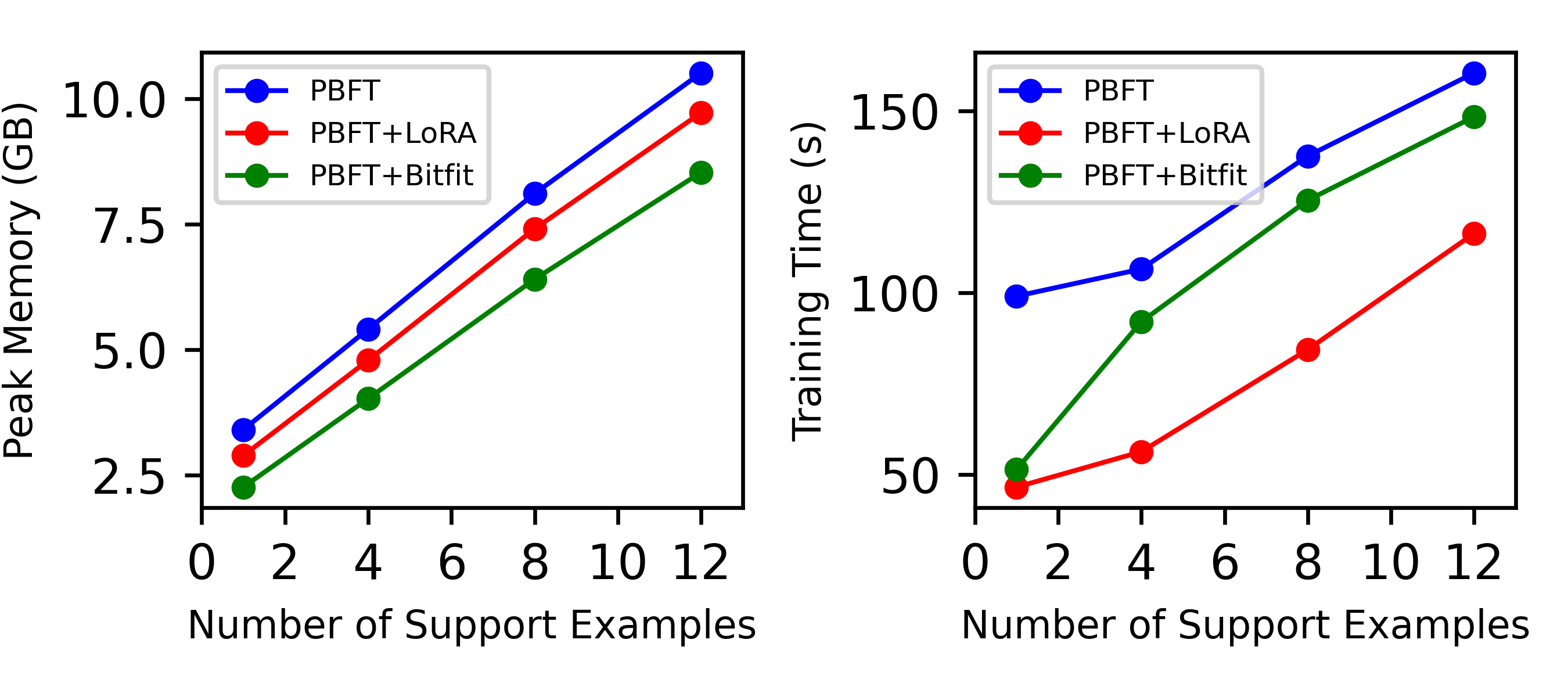}
\end{center}
   \caption{\textcolor{yifei}{Peak GPU memory allocated and training time for PBFT OPT-125M model using LoRA and BitFit.}}
   \label{fig:125m-pbft-eff}
\end{figure}

\textcolor{zirui}{
Memory consumption patterns reveal distinct advantages for both parameter-efficient variants (Figure \ref{fig:125m-pbft-eff}, left panel). Baseline PBFT shows the highest memory requirements, scaling linearly to approximately 10.5GB with 12 support examples. By contrast, both PBFT+BitFit and PBFT+LoRA achieve significant memory savings, with PBFT+BitFit demonstrating the most efficient memory utilization (reaching about 8.5GB at 12 examples) and PBFT+LoRA occupying an intermediate position at nearly (approximately 9.7GB at 12 examples). While all approaches maintain linear scaling with increased support examples, both parameter-efficient variants show notably lower slopes, indicating better memory efficiency without compromising model effectiveness.
}

\textcolor{zirui}{
Training time analysis reveals significant variations among the three approaches (Figure \ref{fig:125m-pbft-eff}, right panel). Baseline PBFT consistently incurs the greatest overhead, requiring approximately 160 seconds with 12 support examples. By contrast, PBFT+LoRA demonstrates the most substantial improvement in training efficiency, reducing the required time to nearly 116 seconds under the same conditions. PBFT+BitFit offers an intermediate improvement, needing roughly 148 seconds. All methods exhibit linear scaling in training time with increased support examples, but PBFT+LoRA maintains a consistently lower slope, suggesting superior scalability for larger datasets.
}

\textcolor{zirui}{
These results highlight the complementary strengths of the parameter-efficient approaches. PBFT+BitFit excels in maintaining stable task performance while reducing memory usage, while PBFT+LoRA achieves the fastest training time efficiency. Thus, the choice between these variants should be guided by specific application requirements: PBFT+BitFit is preferable when task performance and memory constraints are paramount, while PBFT+LoRA is advantageous when training speed is the primary concern. Both variants successfully address the scalability limitations of the baseline PBFT, providing valuable trade-offs to suit diverse operational requirements.
}

\begin{figure}[h]
\begin{center}
   \includegraphics[width=\linewidth]{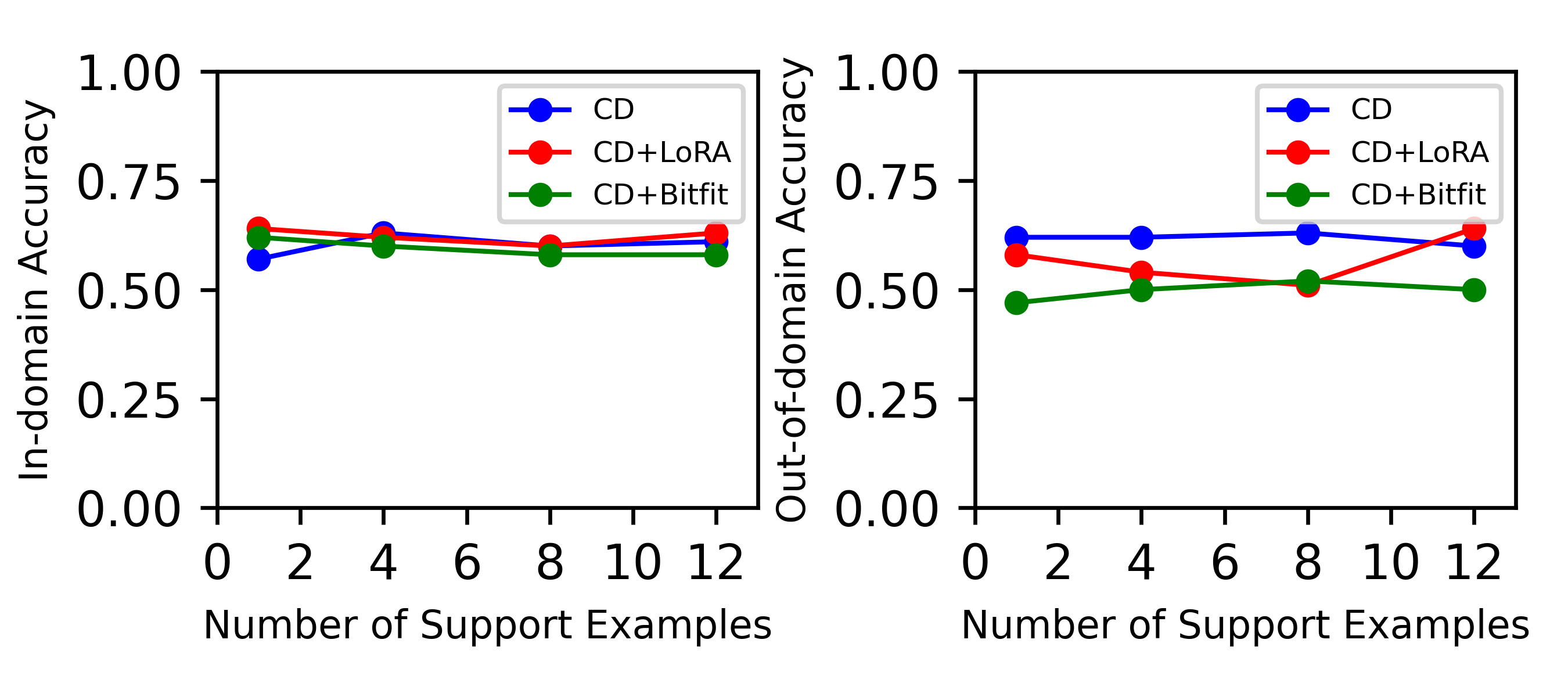}
\end{center}
   \caption{\textcolor{yifei}{In-domain and out-of-domain accuracy for OPT-125M student model distilled from OPT-1.3B teacher model with and without LoRA or BitFit.}}
   \label{fig:cd-acc}
\end{figure}

\textcolor{zirui}{
\subsection{CD+BitFit and CD+LoRA Analysis}
Our experimental results reveal interesting patterns in the performance of Context Distillation (CD) and its parameter-efficient variants (Figure \ref{fig:cd-acc}). For in-domain accuracy, all three approaches demonstrate remarkably stable performance across different numbers of support examples, maintaining accuracies between 0.57-0.66. CD+BitFit slightly outperforms the others at higher support counts, reaching a peak accuracy of 0.66 at 12 supports, while baseline CD and CD+LoRA maintain consistent performance around 0.60-0.63. Notably, unlike the earlier PBFT experiments, none of the variants exhibit the performance degradation as support examples grow, highlighting improved robustness and scalability.
In out-of-domain generalization, the methods show more distinct patterns. Baseline CD maintains the most stable performance at around 0.6-0.63 accuracy across all support examples. CD+LoRA shows variable performance, starting lower at about 0.58 with just 1 support, but steadily improves, ultimately reaching 0.64 at 12 supports, thus demonstrating enhanced adaptability as more examples become available. By constrast, CD+BitFit demonstrates the most conservative performance, maintaining steady but lower accuracy around 0.5 throughout all support example quantities. Although this consistency avoids drastic drops, it lags behind CD and CD+LoRA’s higher out-of-domain scores.
}

\begin{figure}[h]
\begin{center}
   \includegraphics[width=\linewidth]{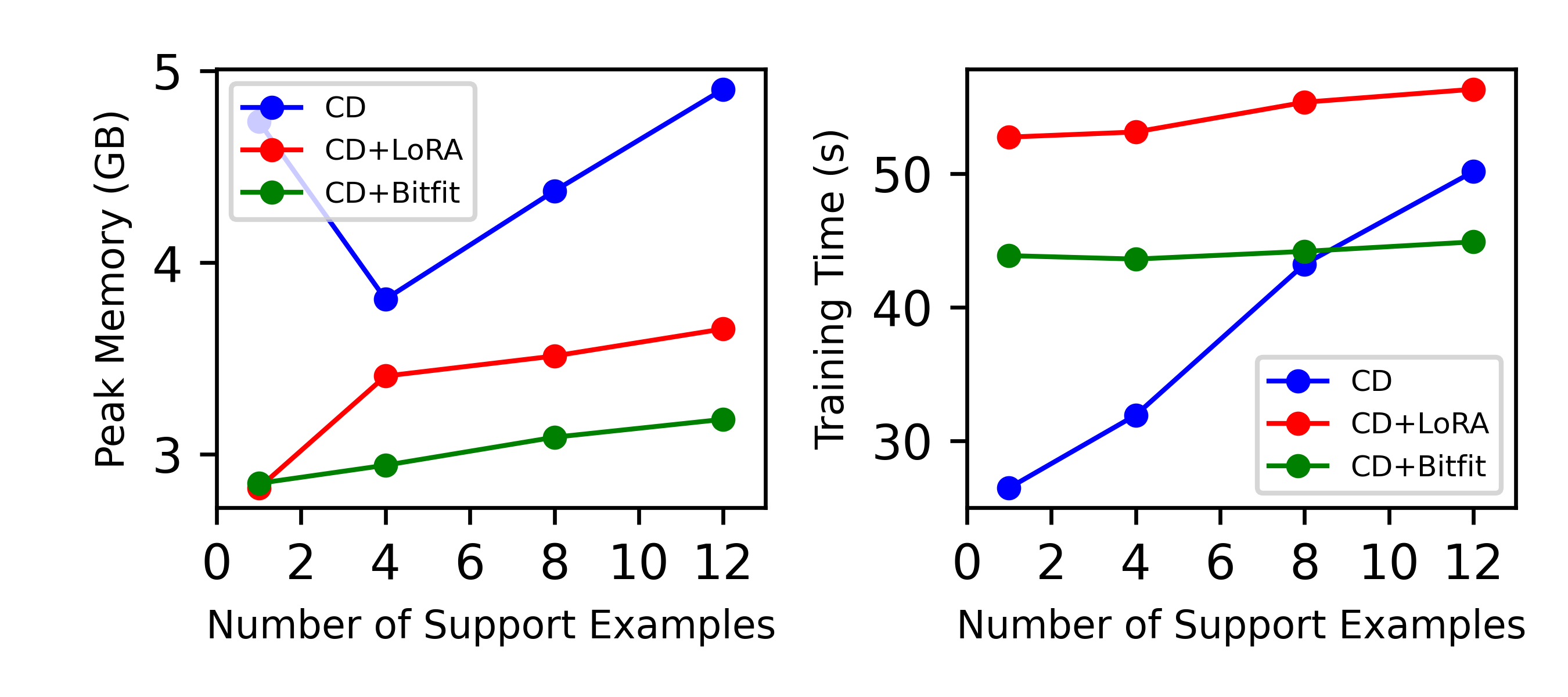}
\end{center}
   \caption{\textcolor{yifei}{Peak GPU memory allocated and training time for applying context-distillation to OPT-125M-student-and-OPT-1.3B-teacher model with and without LoRA or BitFit.}}
   \label{fig:cd-eff}
\end{figure}

\textcolor{zirui}{
Memory consumption patterns reveal significant differences among the three approaches (Figure \ref{fig:cd-eff}, left panel). CD+BitFit shows the highest but most stable memory usage, maintaining around 5.5GB across different numbers of support examples. The baseline CD demonstrates more variable memory consumption, starting at approximately 4.7GB, dropping to 3.8GB at 4 support examples, and then increasing again to about 4.9GB with 12 examples. CD+LoRA achieves the most efficient memory utilization, starting at around 2.8GB and stabilizing at approximately 3.6GB with higher numbers of support examples, showing minimal increase beyond 8 support examples. Such modest increments highlight CD+LoRA’s ability to accommodate larger support sets without incurring steep memory penalties.
}

\textcolor{zirui}{
Training time efficiency varies notably across the three methods (Figure \ref{fig:cd-eff}, right panel). CD+BitFit shows the highest but most consistent training time, requiring around 65 seconds regardless of the number of support examples. CD+LoRA demonstrates intermediate efficiency, maintaining relatively stable training times around 55 seconds. The baseline CD shows the most variable pattern, starting with the lowest training time (approximately 27 seconds) but scaling up more significantly with increased support examples, reaching about 50 seconds with 12 examples.
}

\textcolor{zirui}{
These results suggest that each variant offers distinct advantages: CD+LoRA provides the best balance of memory efficiency and training time stability, while maintaining competitive task performance. CD+BitFit offers the most stable (though higher) resource utilization and slightly better in-domain performance with fewer examples. The baseline CD shows superior out-of-domain generalization but at the cost of less predictable resource utilization. These trade-offs suggest that the choice between variants should be guided by specific application requirements and resource constraints.
}

\section{Discussion}

\subsection{Summary}

\textcolor{yifei}{
Context distillation is employed to fine-tune the OPT-125M model using an OPT-1.3B teacher model, demonstrating significantly improved knowledge transfer with nearly 50\% higher out-of-domain accuracy compared to in-context learning. Additionally, context distillation has substantially lower hardware requirements than conventional pattern-based fine-tuning, reducing peak memory usage by up to 60\%. 
On top of that, the memory requirements for context distillation remain nearly independent of context size, unlike pattern-based fine-tuning, where memory usage increases nonlinearly with context size. 
By enabling few-shot fine-tuning without consuming input token limits for context, context distillation ensures that the context window is dedicated to the query prompt, accelerating the inference process. This approach also eliminates constraints on context size, making it suitable for handling long conversations, entire documents, or extensive code bases.
}

\subsection{Limitations and Future Work}
\textcolor{liu}{
This study has several limitations that provide opportunities for future research. Hardware constraints restricted fine-tuning experiments to small-scale open-source models (OPT-125M, 350M, and 1.3B), with larger models like OPT-6.7B facing out-of-memory errors on A100 GPUs. Time constraints limited hyperparameter exploration, hindering deeper insights into model optimization. The study focused on a single task, leaving the generalizability of Context Distillation across diverse tasks unexplored. Minimal effort was invested in prompt engineering, which is critical for model performance, warranting future studies to design and evaluate more effective prompts. Additionally, challenges with mixed-precision training impacted memory efficiency; resolving these issues could enable experiments with larger models and batch sizes. Addressing these limitations will be crucial for advancing Context Distillation research.
}

\bibliographystyle{ieeetr}
\bibliography{paper}

\end{document}